\documentclass[11pt,a4paper]{article}
\usepackage[hyperref]{acl2020}
\usepackage{times}
\usepackage{latexsym}

\usepackage{microtype}

\usepackage{url}
\usepackage{graphicx}
\usepackage{wrapfig}
\usepackage{amsmath}
\usepackage{amssymb}
\usepackage{color,xcolor,colortbl}
\usepackage{enumitem}
\usepackage{algorithm}
\usepackage{algorithmic}
\usepackage{bm,bbm}
\usepackage{booktabs}
\usepackage{mathtools}
\usepackage{array}
\usepackage{multirow}
\usepackage{soul}
\usepackage{subcaption}
\usepackage{pbox}
\usepackage{pifont}
\usepackage{arydshln}

\newcommand\Tstrut{\rule{0pt}{2.6ex}}         

\definecolor{darkblue}{rgb}{0.0, 0.0, 0.55}
\setul{0.5ex}{0.3ex} 
\setulcolor{blue} 
\aclfinalcopy 
\def\aclpaperid{109} 

\title{Understanding Points of Correspondence between Sentences for Abstractive Summarization}

\author{Logan Lebanoff$^\dagger$ \;\; John Muchovej$^\dagger$ \;\; Franck Dernoncourt$^\S$ \\
\textbf{Doo Soon Kim$^\S$ \quad\; Lidan Wang$^\S$ \quad\; Walter Chang$^\S$ \quad\; Fei Liu$^\dagger$}
\\[0.8em]
$^\dagger$University of Central Florida \quad $^\S$Adobe Research\\
\texttt{\small \{loganlebanoff,john.muchovej\}@knights.ucf.edu} \quad \texttt{\small feiliu@cs.ucf.edu}\\ 
\texttt{\small \{dernonco,dkim,lidwang,wachang\}@adobe.com}
}

\date{}

\begin{document}

\maketitle
\begin{abstract}

Fusing sentences containing disparate content is a remarkable human ability that helps create informative and succinct summaries. 
Such a simple task for humans has remained challenging for modern abstractive summarizers, substantially restricting their applicability in real-world scenarios. 
In this paper, we present an investigation into fusing sentences drawn from a document by introducing the notion of points of correspondence, which are cohesive devices that tie any two sentences together into a coherent text.
The types of points of correspondence are delineated by text cohesion theory, covering pronominal and nominal referencing, repetition and beyond. 
We create a dataset containing the documents, source and fusion sentences, and human annotations of points of correspondence between sentences.
Our dataset bridges the gap between coreference resolution and summarization.
It is publicly shared to serve as a basis for future work to measure the success of sentence fusion systems.\footnote{\footnotesize\url{https://github.com/ucfnlp/points-of-correspondence}}

\end{abstract}

\section{Introduction}
\label{sec:intro}

\begin{table}[t]
\setlength{\tabcolsep}{5pt}
\renewcommand{\arraystretch}{1.1}
\centering
\begin{scriptsize}
\textsf{
\begin{tabular}{|l|}
\hline
\textbf{[Source Sentences]}\Tstrut\\[0.5em]
\textcolor{blue}{Robert Downey Jr.} is making headlines for walking out of an\\
interview with a \textcolor{black}{British journalist} who dared to veer away from the\\
superhero movie Downey was there to promote.\\[0.5em]
\textcolor{red}{The journalist instead started asking personal questions about the}\\ 
\textcolor{red}{actor's political beliefs} and ``dark periods'' of addiction and jail time.\\[0.5em]
\hline
\hline
\textbf{[Summary]}\, \textcolor{blue}{Robert Downey Jr} \textcolor{red}{started asking personal questions}\Tstrut\\
\textcolor{red}{about the actor's political beliefs.}\\[0.2em]
\hline
\hline
\textbf{[Source Sentences]}\Tstrut\\[0.5em]
\textcolor{black}{``Real Housewives of Beverly Hills'' star and former child actress}\\
\textcolor{blue}{Kim Richards} \textcolor{blue}{is accused of kicking} \textcolor{blue}{a police officer} after being\\
arrested Thursday morning.\\[0.5em]
\textcolor{red}{A police representative} \textcolor{red}{said} \textcolor{red}{Richards} \textcolor{red}{was asked to leave but}\\
\textcolor{red}{refused} and then entered a restroom and wouldn't come out. \\[0.5em]
\hline
\hline
\textbf{[Summary]}\, \textcolor{blue}{Kim Richards} \textcolor{blue}{is accused of kicking} \textcolor{blue}{a police officer}\Tstrut\\
\textcolor{red}{who refused to leave.}\\[0.2em]
\hline
\hline
\textbf{[Source Sentences]}\Tstrut\\[0.5em]
The kind of horror represented by the Blackwater case and others\\
like it [...] may be largely absent from public memory in the West\\ 
these days, but it is being \textcolor{red}{used} \textcolor{red}{by the Islamic State in Iraq and}\\
\textcolor{red}{Syria (ISIS)} to support its sectarian narrative.\\[0.5em]
\textcolor{blue}{In its propaganda, ISIS has been} \textcolor{blue}{using} Abu Ghraib and other\\
cases of Western abuse to legitimize its current actions [...]\\[0.5em]
\hline
\hline
\textbf{[Summary]}\, \textcolor{blue}{In its propaganda, ISIS is being} \textcolor{red}{used by the Islamic}\Tstrut\\
\textcolor{red}{State in Iraq and Syria.}\\[0.2em]
\hline
\end{tabular}}
\end{scriptsize}
\vspace{-0.075in}
\caption{
Unfaithful summary sentences generated by neural abstractive summarizers, in-house and PG~\cite{See:2017}.
They attempt to merge two sentences into one sentence with improper use of \emph{points of correspondence} between sentences, yielding nonsensical output.
Summaries are manually re-cased for readability.}
\label{tab:example}
\vspace{-0.1in}
\end{table}

\begin{table*}
\setlength{\tabcolsep}{3pt}
\renewcommand{\arraystretch}{1.2}
\centering
\begin{scriptsize}
\textsf{
\begin{tabular}{|l|l|l|}
\hline
\textbf{PoC Type}\quad\quad\quad\quad & \textbf{Source Sentences} & \textbf{Summary Sentence} \\
\hline
\hline
\textbf{Pronominal}  & [\textbf{S1}] \colorbox{red!30}{The bodies} showed signs of torture. & $\bullet$ \colorbox{red!30}{The bodies} of the men, which showed signs \\
Referencing & [\textbf{S2}] \colorbox{red!30}{They} were left on the side of a highway in Chilpancingo, about an  & of torture, were left on the side of a highway \\
& hour north of the tourist resort of Acapulco in the state of Guerrero. & in Chilpancingo.\\
\hline
\textbf{Nominal} & [\textbf{S1}] \colorbox{orange!30}{Bahamian R\&B singer Johnny Kemp}, best known for the 1988 party & $\bullet$ \colorbox{orange!30}{Johnny Kemp} is ``believed to have \colorbox{gray!0}{drowned} at \\
Referencing & anthem “Just Got Paid,” \colorbox{gray!0}{died} this week in Jamaica. & a beach in Montego Bay,'' police say. \\
& [\textbf{S2}] \colorbox{orange!30}{The singer} is believed to have \colorbox{gray!0}{drowned} at a beach in Montego Bay & \\
& on Thursday, the Jamaica Constabulatory Force said in a press release. & \\
\hline
\textbf{Common-Noun} & [\textbf{S1}] A nurse confessed to killing \colorbox{magenta!30}{five women and one man} at hospital.  & $\bullet$ The nurse, who has been dubbed “nurse \\
Referencing & [\textbf{S2}] A former nurse in the Czech Republic murdered \colorbox{magenta!30}{six of her elderly} & death” locally, has admitted killing \colorbox{magenta!30}{the victims} \\
& \colorbox{magenta!30}{patients} with massive doses of potassium in order to ease her workload. & with massive doses of potassium.\\
\hline
\textbf{Repetition} & [\textbf{S1}] Stewart said that she and her husband, Joseph Naaman, booked & $\bullet$ Couple spends \$1,200 to ship their cat, \colorbox{blue!30}{Felix},  \\
& \colorbox{blue!30}{Felix} on their flight from the United Arab Emirates to New York on April 1. & on a flight from the United Arab Emirates.\\
& [\textbf{S2}] The couple said they spent \$1,200 to ship \colorbox{blue!30}{Felix} on the 14-hour flight. & \\
\hline
\textbf{Event Triggers} & [\textbf{S1}] \colorbox{gray!0}{Four employees of the store} have been \colorbox{lime!30}{arrested}, but its manager & $\bullet$ \colorbox{gray!0}{The four store workers} \colorbox{lime!30}{arrested} could spend \\
& was still at large, said Goa police superintendent Kartik Kashyap. & 3 years each in prison if \colorbox{lime!30}{convicted}.\\
& [\textbf{S2}] If \colorbox{lime!30}{convicted}, \colorbox{gray!0}{they} could spend up to three years in jail, Kashyap said. & \\
\hline
\end{tabular}
}
\end{scriptsize}
\vspace{-0.05in}
\caption{Types of sentence correspondences.
Text cohesion can manifest itself in different forms.
}
\label{tab:poc_type}
\vspace{-0.1in}
\end{table*}

Stitching portions of text together into a sentence is a crucial first step in abstractive summarization.
It involves choosing which sentences to fuse, what content from each of them to retain and how best to present that information~\cite{Elsner:2011}.
A major challenge in fusing sentences is to establish correspondence between sentences.
If there exists no correspondence, it would be difficult, if not impossible, to fuse sentences.
In Table~\ref{tab:example}, we present example source and fusion sentences, where the summarizer attempts to merge two sentences into a summary sentence with improper use of \emph{points of correspondence}.
In this paper, we seek to uncover hidden correspondences between sentences, which has a great potential for improving content selection and deep sentence fusion.

Sentence fusion (or multi-sentence compression) plays a prominent role in automated summarization and its importance has long been recognized~\cite{Barzilay:1999}. 
Early attempts to fuse sentences build a dependency graph from sentences, then decode a tree from the graph using integer linear programming, finally linearize the tree to generate a summary sentence~\cite{Barzilay:2005,Filippova:2008,Thadani:2013}.
Despite valuable insights gained from these attempts, experiments are often performed on small datasets and systems are designed to merge sentences conveying \emph{similar} information. 
Nonetheless, humans do not restrict themselves to combine similar sentences, but also \emph{disparate} sentences containing fundamentally different content but remain related to make fusion sensible~\cite{Elsner:2011}.
We focus specifically on analyzing fusion of \textit{disparate} sentences, which is a distinct problem from fusing a set of \textit{similar} sentences.

While fusing disparate sentences is a seemingly simple task for humans to do, it has remained challenging for modern abstractive summarizers~\cite{See:2017,Celikyilmaz:2018,Chen:2018:ACL,Liu:2019:HTrans}.
These systems learn to perform content selection and generation through end-to-end learning.
However, such a strategy is not consistently effective and they struggle to reliably perform sentence fusion~\cite{Falke:2019,Kryscinski:2019}.
E.g., only 6\% of summary sentences generated by pointer-generator networks~\cite{See:2017} are fusion sentences; the ratio for human abstracts is much higher (32\%). 
Further, Lebanoff et al.~\shortcite{Lebanoff:2019:WS} report that 38\% of fusion sentences contain incorrect facts.
There exists a pressing need for---and this paper contributes to--broadening the understanding of points of correspondence used for sentence fusion.

We present the first attempt to construct a sizeable sentence fusion dataset, where an instance in the dataset consists of a pair of input sentences, a fusion sentence, and human-annotated \emph{points of correspondence} between sentences. 
Distinguishing our work from previous efforts~\cite{Geva:2019}, our input contains \emph{disparate} sentences and output is a fusion sentence containing important, though not equivalent information of the input sentences. 
Our investigation is inspired by Halliday and Hasan's theory of \emph{text cohesion}~\shortcite{Halliday:1976} that covers a broad range of points of correspondence, including entity and event coreference~\cite{Ng:2017,Lu:2018}, shared words/concepts between sentences and more.
Our contributions are as follows.
\begin{itemize}[topsep=5pt,itemsep=0pt,leftmargin=*]
\item We describe the first effort at establishing points of correspondence between disparate sentences.
Without a clear understanding of points of correspondence, sentence fusion remains a daunting challenge that is only sparsely and sometimes incorrectly performed by abstractive summarizers.

\item We present a sizable dataset for sentence fusion containing human-annotated corresponding regions between pairs of sentences. 
It can be used as a testbed for evaluating the ability of summarization models to perform sentence fusion.
We report on the insights gained from annotations to suggest important future directions for sentence fusion. 
Our dataset is released publicly.

\end{itemize}

\section{Annotating Points of Correspondence}
\label{sec:annotation}

We cast sentence fusion as a constrained summarization task where portions of text are selected from each source sentence and stitched together to form a fusion sentence; rephrasing and reordering are allowed in this process.
We propose guidelines for annotating \emph{points of correspondence} (PoC) between sentences based on Halliday and Hasan's theory of cohesion~\shortcite{Halliday:1976}.

We consider {points of correspondence} as cohesive phrases that tie sentences together into a coherent text.
Guided by text cohesion theory, we categorize PoC into five types, including {pronominal referencing} (``\emph{they}''), {nominal referencing} (``\emph{Johnny Kemp}''), {common-noun referencing} (``\emph{five women}''), {repetition}, and {event trigger words} that are related in meaning (``\emph{died}'' and ``\emph{drowned}'').
An illustration of PoC types is provided in Table~\ref{tab:poc_type}.
Our categorization emphasizes the lexical linking that holds a text together and gives it meaning.

\begin{figure}
\centering
\includegraphics[width=3in]{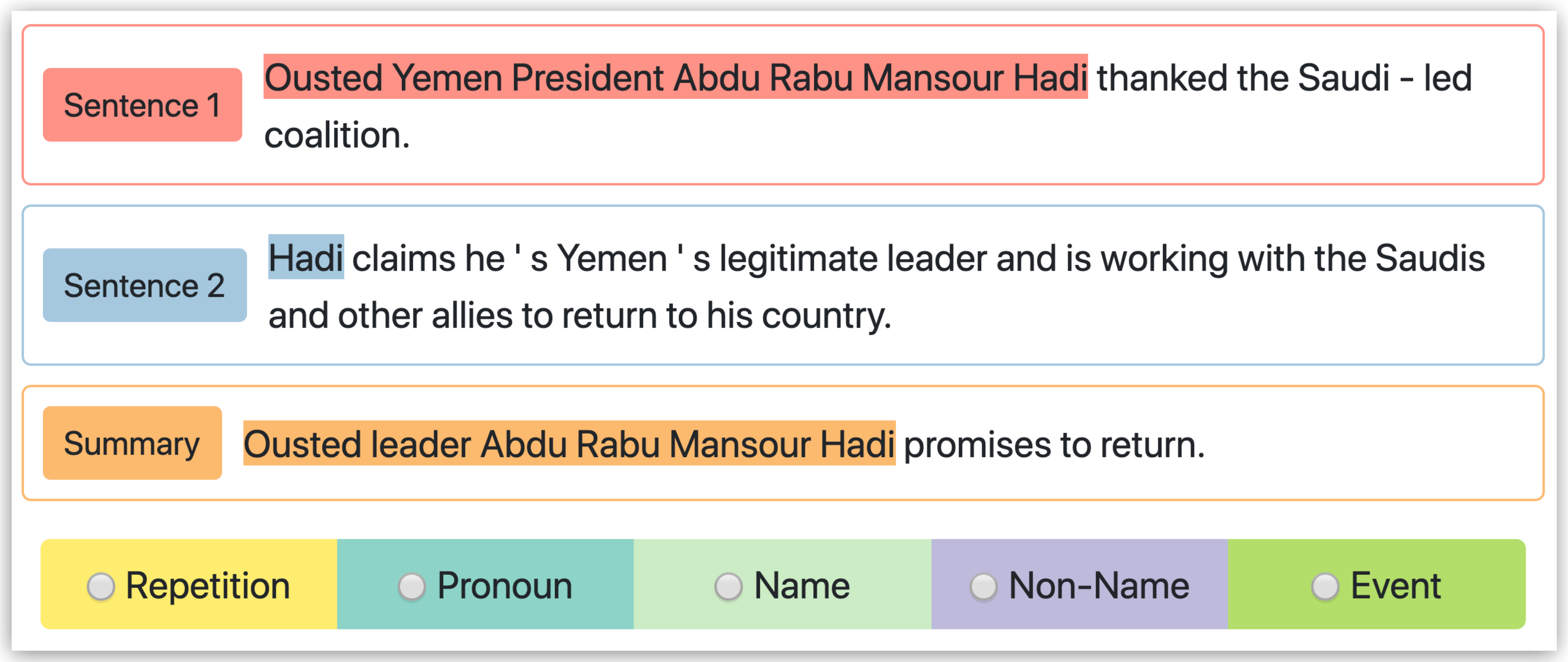}
\caption{An illustration of the annotation interface.
A human annotator is asked to highlight text spans referring to the same entity, then choose one from the five pre-defined PoC types.
}
\label{fig:poc_anno}
\vspace{-0.1in}
\end{figure}

A human annotator is instructed to identify a text span from each of the source sentences and summary sentence, thus establishing a point of correspondence between source sentences, and between source and summary sentences.
As our goal is to understand the role of PoC in sentence fusion, we do not consider the case if PoC is only found in source sentences but not summary sentence, e.g., ``\emph{Kashyap said}'' and ``\emph{said Goa police superintendent Kartik Kashyap}'' in Table~\ref{tab:poc_type}.
If multiple PoC co-exist in an example, an annotator is expected to label them all; a separate PoC type will be assigned to each PoC occurrence.
We are particularly interested in annotating inter-sentence PoC. 
If entity mentions (``\emph{John}'' and ``\emph{he}'') are found in the same sentence, we do not explicitly label them but assume such intra-sentence referencing can be captured by an existing coreference resolver.
Instances of source sentences and summary sentences are obtained from the test and validation splits of the CNN/DailyMail corpus~\cite{See:2017} following the procedure described by Lebanoff et al.~\shortcite{Lebanoff:2019:WS}.
We take a human summary sentence as an anchor point to find two document sentences that are most similar to it based on ROUGE. It becomes an instance containing a pair of source sentences and their summary.
The method allows us to identify a large quantity of candidate fusion instances.

Annotations are performed in two stages.  
Stage one removes all spurious pairs that are generated by the heuristic, i.e. a summary sentence that is not a valid fusion of the corresponding two source sentences.
Human annotators are given a pair of sentences and a summary sentence and are asked whether it represents a valid fusion.
The pairs identified as valid fusions by a majority of annotators move on to stage two.
Stage two identifies the corresponding regions in the sentences.
As shown in Figure~\ref{fig:poc_anno}, annotators are given a pair of sentences and their summary and are tasked with highlighting the corresponding regions between each sentence.
They must also choose one of the five PoC types (repetition, pronominal, nominal, common-noun referencing, and event triggers) for the set of corresponding regions.

\begin{figure}[t]
\centering
\includegraphics[width=0.45\textwidth]{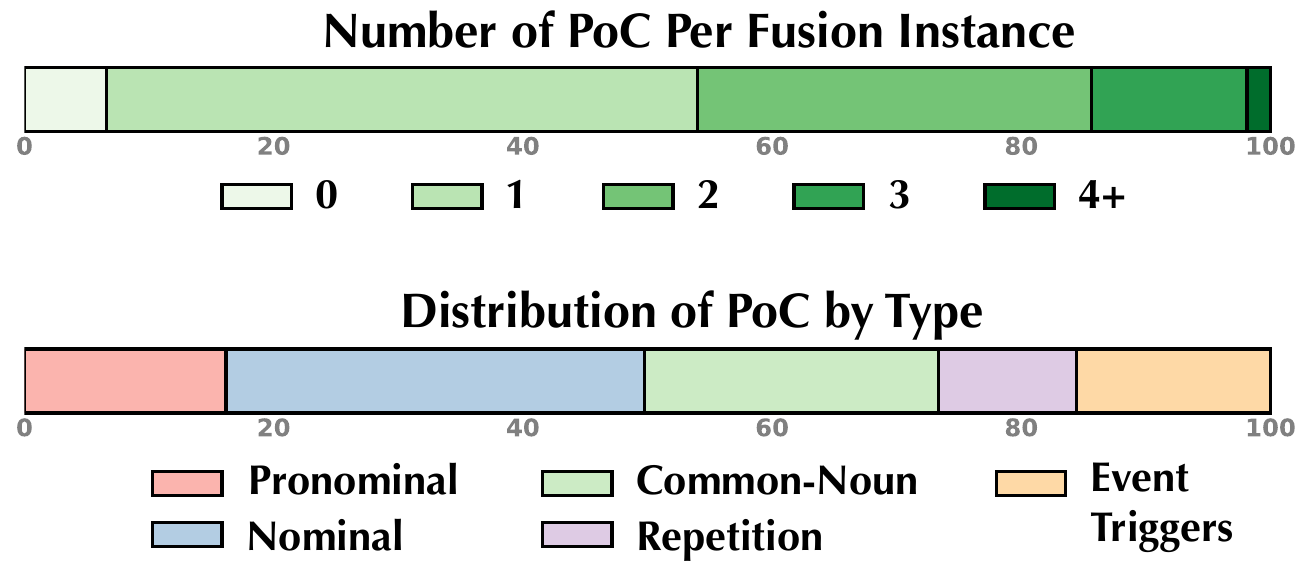}
\vspace{-0.05in}
\caption{Statistics of PoC occurrences and types.}
\label{fig:statistics}
\vspace{-0.1in}
\end{figure}

\begin{table*}
\setlength{\tabcolsep}{6.5pt}
\renewcommand{\arraystretch}{1}
\begin{center}{\footnotesize
\begin{tabular}{lcccccccc}
\toprule
\textbf{Coref Resolver} & \textbf{P}(\%) & \textbf{R}(\%) &\textbf{F}(\%) & \textbf{Pronominal} & \textbf{Nominal} &\textbf{Comm.-Noun}  & \textbf{Repetition} & \textbf{Event Trig.} \\
\midrule
SpaCy & \textbf{59.2} & 20.1 & 30.0 & 30.8 & 23.3 & 10.4 & 39.9 & 2.6\\
AllenNLP & 49.0 & 24.5 & 32.7 & 36.5 & 28.1 & 14.7 & 47.1 &3.1\\
Stanford CoreNLP & 54.2 & \textbf{26.2} & \textbf{35.3} & 40.0 & 27.3 & 17.4 & 55.1 & 2.3\\
\bottomrule
\end{tabular}}
\end{center}
\vspace{-0.1in}
\caption{
Results of various coreference resolvers on successfully identifying inter-sentence points of correspondence (PoC) and recall scores of these resolvers split by PoC correspondence type.
}
\label{tab:results} 
\end{table*}

We use Amazon mechanical turk, allowing only workers with 95\% approval rate and at least 5,000 accepted tasks.
To ensure high quality annotations, we first run a qualification round of 10 tasks. Workers performing sufficiently on these tasks were allowed to annotate the whole dataset.
For task one, 2,200 instances were evaluated and 621 of them were filtered out.
In total, we annotate points of correspondence for \textbf{1,599 instances, taken from 1,174 documents}.
Similar to \cite{hardy2019highres}, we report Fleiss' Kappa judged on each word (highlighted or not), yielding substantial inter-annotator agreement ($\kappa$=0.58) for annotating points of correspondence.
We include a reference to the original article that each instance was taken from, thus providing context for each instance.

Figure \ref{fig:statistics} shows statistics of PoC occurrence frequencies and the distribution of different PoC types.
A majority of sentence pairs have one or two points of correspondence. 
Only a small percentage (6.5\%) do not share a PoC. A qualitatively analysis shows that these sentences often have an \textit{implicit} discourse relationship, e.g., ``\emph{The two men speak. Scott then gets out of the car, again, and runs away.}'' In this example, there is no clear portion of text that is shared between the sentences; rather, the connection lies in the fact that one event happens after the other.
Most of the PoC are a flavor of coreference (pronominal, nominal, or common-noun). Few are exact repetition. Further, we find that only 38\% of points of correspondence in the sentence pair share any words (lemmatized). This makes identifying them automatically challenging, requiring a deeper understanding of what connects the two sentences.

\section{Resolving Coreference}

Coreference resolution~\cite{Ng:2017} is similar to the task of identifying points of correspondence.
Thus, a natural question we ask is how well state-of-the-art coreference resolvers can be adapted to this task.
If coreference resolvers can perform reasonably well on PoC identification, then these resolvers can be used to extract PoC annotations to potentially enhance sentence fusion. 
If they perform poorly, coreference performance results can indicate areas of improvement for future work on detecting points of correspondence. 
In this paper, we compare three coreference resolvers on our dataset, provided by open-source libraries: Stanford CoreNLP \cite{manning-EtAl:2014:P14-5}, SpaCy \cite{spacy2}, and AllenNLP \cite{Gardner2017AllenNLP}.

We base our evaluation on the standard metric used for coreference resolution, B-CUBED algorithm \cite{bagga1998algorithms}, with some modifications. Each resolver is run on an input pair of sentences to obtain multiple clusters, each representing an entity (e.g., \textit{Johnny Kemp}) containing multiple mentions (e.g., \textit{Johnny Kemp}; \textit{he}; \textit{the singer}) of that entity. More than one cluster can be detected by the coreference resolver, as additional entities may exist in the given sentence pair (e.g., \textit{Johnny Kemp} and \textit{the police}). Similarly, in Section \S\ref{sec:annotation}, human annotators identified multiple PoC clusters, each representing a point of correspondence containing one mention from each sentence. We evaluate how well the resolver-detected clusters compare to the human-detected clusters (i.e., PoCs). If a resolver cluster overlaps both mentions for the gold-standard PoC, then this resolver cluster is classified as a hit. Any resolver cluster that does not overlap both PoC mentions is a miss. Using this metric, we can calculate precision, recall, and F1 scores based on correctly/incorrectly identified tokens from the outputs of each resolver.

The results are presented in Table \ref{tab:results}. The three resolvers exhibit similar performance, but the scores on identifying points of correspondence are less than satisfying.
The SpaCy resolver has the highest precision (59.2\%) and Stanford CoreNLP achieves the highest F1-score (35.3\%).
We observe that existing coreference resolvers can sometimes struggle to use the high-level reasoning that humans use to determine what connects two sentences together. 
Next, we go deeper into understanding what PoC types these resolvers struggle with. 
We present the recall scores of these resolvers split by PoC correspondence type. 
Event coreference poses the most difficulty by far, which is understandable as coreference resolution only focuses on entities rather than events. 
More work into detecting event coreference can bring significant improvements in PoC identification. 
Common-noun coreference also poses a challenge, in part because names and pronouns give strong clues as to the relationships between mentions, while common-noun relationships are more difficult to identify since they lack these clues.

\section{Sentence Fusion}
\label{sec:fusion}

Truly effective summarization will only be achievable when systems have the ability to fully recognize points of correspondence between sentences.
It remains to be seen whether such knowledge can be acquired implicitly by neural abstractive systems through joint content selection and generation.  
We next conduct an initial study to assess neural abstractive summarizers on their ability to perform sentence fusion to merge two sentences into a summary sentence.
The task represents an important, atomic unit of abstractive summarization, because a long summary is still generated one sentence at a time~\cite{Lebanoff:2019}.

We compare two best-performing abstractive summarizers: 
\textit{Pointer-Generator} uses an encoder-decoder architecture with attention and copy mechanism~\cite{See:2017}; 
\textit{Transformer} adopts a decoder-only Transformer architecture similar to that of~\cite{Radford:2019}, where a summary is decoded one word at a time conditioned on source sentences and the previously-generated summary words.
We use the same number of heads, layers, and units per layer as BERT-base \cite{Devlin:2018}.
In both cases, the summarizer was trained on about 100k instances derived from the train split of CNN/DailyMail, using the same heuristic as described in (\S\ref{sec:annotation}) without PoC annotations.
The summarizer is then tested on our dataset of 1,599 fusion instances and evaluated using standard metrics~\cite{Lin:2004}. We also report how often each summarizer actually draws content from both sentences (\textit{\%Fuse}), rather than taking content from only one sentence. A generated sentence counts as a fusion if it contains at least two non-stopword tokens from each sentence not already present in the other sentence.
Additionally, we include a \textit{Concat-Baseline} creating a fusion sentence by simply concatenating the two source sentences.

\begin{table}[t]
\setlength{\tabcolsep}{6.5pt}
\renewcommand{\arraystretch}{1.15}
\begin{center}{\footnotesize
\begin{tabular}{ l c c c r}
\toprule
\textbf{System} & \textbf{R-1} & \textbf{R-2} &\textbf{R-L} & \textbf{\%Fuse}\\
\midrule
Concat-Baseline & 36.13 & 18.64 & 27.79 & 99.7 \\
Pointer-Generator & 33.74 & 16.32 & 29.27 & 38.7\\
Transformer & \textbf{38.81} & \textbf{20.03} & \textbf{33.79} & 50.7 \\
\bottomrule
\end{tabular}}
\end{center}
\vspace{-0.1in}
\caption{ROUGE scores of neural abstractive summarizers on the sentence fusion dataset.
We also report the percentage of output sentences that are indeed fusion sentences (\%Fuse) }
\label{tab:fusion_results} 
\vspace{-0.1in}
\end{table}

The results according to the ROUGE evaluation \cite{Lin:2004} are presented in Table \ref{tab:fusion_results}. 
Sentence fusion appears to be a challenging task even for modern abstractive summarizers.
Pointer-Generator has been shown to perform strongly on abstractive summarization, but it is less so on sentence fusion and in other highly abstractive settings~\cite{narayan2018dont}.
Transformer significantly outperforms other methods, in line with previous findings~\cite{Liu2018wikipedia}. 
We qualitatively examine system outputs.
Table~\ref{tab:example} presents fusions generated by these models and exemplifies the need for infusing models with knowledge of points of correspondence. In the first example, Pointer-Generator incorrectly conflates \textit{Robert Downey Jr.} with the \textit{journalist} asking questions. Similarly, in the second example, Transformer states the \textit{police officer} refused to leave when it was actually \textit{Richards}. Had the models explicitly recognized the points of correspondence in the sentences---that \textit{the journalist} is a separate entity from \textit{Robert Downey Jr.} and that \textit{Richards} is separate from \textit{police officer}---then a more accurate summary could have been generated.

\section{Related Work}
\label{sec:related}

Uncovering hidden correspondences between sentences is essential for producing proper summary sentences.
A number of recent efforts select important words and sentences from a given document, then let the summarizer attend to selected content to generate a summary~\cite{Gehrmann:2018,Hsu:2018,Chen:2018:ACL,Putra2018,Lebanoff:2018,Liu:2019:HTrans}.
These systems are largely agnostic to sentence correspondences, which can have two undesirable consequences.
If only a single sentence is selected, it can be impossible for the summarizer to produce a fusion sentence from it.
Moreover, if \emph{non-fusible} textual units are selected, the summarizer is forced to fuse them into a summary sentence, yielding output summaries that often fail to keep the original meaning intact.
Therefore, in this paper we had investigated the correspondences between sentences to gain an understanding of sentence fusion.

Establishing correspondence between sentences goes beyond finding common words.
Humans can fuse sentences sharing \emph{few or no} common words if they can find other types of correspondence.
Fusing such disparate sentences poses a serious challenge for automated fusion systems~\cite{Marsi:2005,Filippova:2008,McKeown:2010,Elsner:2011,Thadani:2013:IJCNLP,Mehdad:2013,Nayeem:2018}. 
These systems rely on common words to derive a connected graph from input sentences or subject-verb-object triples~\cite{Moryossef:2019}.
When there are no common words in sentences, systems tend to break apart.

There has been a lack of annotated datasets and guidelines for sentence fusion. 
Few studies have investigated the types of correspondence between sentences such as entity and event coreference.
Evaluating sentence fusion systems requires not only novel metrics~\cite{zhao-etal-2019-moverscore,Zhang2020BERTScore,durmus2020feqa,wang2020asking} but also high-quality ground-truth annotations.
It is therefore necessary to conduct a first study to look into cues humans use to establish correspondence between disparate sentences.

\begin{table}[t]
\setlength{\tabcolsep}{0pt}
\renewcommand{\arraystretch}{1.1}
\centering
\begin{scriptsize}
\textsf{
\begin{tabular}{p{3in}}
\toprule
\textbf{\cite{McKeown:2010}}\Tstrut\\[0.5em]
\textbf{[S1]} Palin actually turned against the bridge project only after it
became a national symbol of wasteful spending.\\
\textbf{[S2]} Ms. Palin supported the bridge project while running for
governor, and abandoned it after it became a national scandal.\\
\textbf{[Fusion]} Palin turned against the bridge project after it
became a national scandal.\\
\midrule
\textbf{DiscoFuse \cite{Geva:2019}}\Tstrut\\[0.5em]
\textbf{[S1]} Melvyn Douglas originally was signed to play Sam Bailey.\\
\textbf{[S2]} The role ultimately went to Walter Pidgeon.\\
\textbf{[Fusion]} Melvyn Douglas originally was signed to play Sam Bailey, but the role ultimately went to Walter Pidgeon.\\
\midrule
\textbf{Points of Correspondence Dataset (Our Work)}\Tstrut\\[0.5em]
\textbf{[S1]} \colorbox{red!30}{The bodies} showed signs of torture.\\
\textbf{[S2]} \colorbox{red!30}{They} were left on the side of a highway in Chilpancingo, about an  hour north of the tourist resort of Acapulco in the state of Guerrero.\\
\textbf{[Fusion]} \colorbox{red!30}{The bodies} of the men, which showed signs of torture, were left on the side of a highway in Chilpancingo.\\
\bottomrule
\end{tabular}}
\end{scriptsize}
\vspace{-0.05in}
\caption{Comparison of sentence fusion datasets.}
\label{tab:fusion_examples}
\vspace{-0.1in}
\end{table}

We envision sentence correspondence to be related to text \emph{cohesion} and \emph{coherence}, which help establish correspondences between two pieces of text.
Halliday and Hasan \shortcite{Halliday:1976} describe text \textbf{cohesion} as cohesive devices that tie two textual elements together.
They identify five categories of cohesion: \emph{reference}, \emph{lexical cohesion}, \emph{ellipsis}, \emph{substitution} and \emph{conjunction}.
In contrast, \textbf{coherence} is defined in terms of discourse relations between textual elements, such as \emph{elaboration}, \emph{cause} or \emph{explanation}.
Previous work studied discourse relations~\cite{Geva:2019}, this paper instead focuses on \emph{text cohesion}, which plays a crucial role in generating proper fusion sentences. 
Our dataset contains pairs of source and fusion sentences collected from news editors in a natural environment.
The work is particularly meaningful to text-to-text and data-to-text generation~\cite{Gatt:2018} that demand robust modules to merge disparate content.

We contrast our dataset with previous sentence fusion datasets. McKeown et al. \shortcite{McKeown:2010} compile a corpus of 300 sentence fusions as a first step toward a supervised fusion system. However, the input sentences have very similar meaning, though they often present lexical variations and different details. In contrast, our proposed dataset seeks to fuse significantly different meanings together into a single sentence. A large-scale dataset of sentence fusions has been recently collected \cite{Geva:2019}, where each sentence has disparate content and are connected by various discourse connectives. This paper instead focuses on \emph{text cohesion} and on fusing only the salient information, which are both vital for abstractive summarization. Examples are presented in Table~\ref{tab:fusion_examples}.

\section{Conclusion}
\label{sec:conclusion}

In this paper, we describe a first effort at annotating points of correspondence between disparate sentences.
We present a benchmark dataset comprised of the documents, source and fusion sentences, and human annotations of points of correspondence between sentences.
The dataset fills a notable gap of coreference resolution and summarization research.
Our findings shed light on the importance of modeling points of correspondence, suggesting important future directions for sentence fusion.

\section*{Acknowledgments}

We are grateful to the anonymous reviewers for their helpful comments and suggestions.
This research was supported in part by the National Science Foundation grant IIS-1909603.

\bibliography{summ,abs_summ,fei,logan}
\bibliographystyle{acl_natbib}

\end{document}